\documentclass[conference]{IEEEtran}
\IEEEoverridecommandlockouts
\usepackage{cite}
\usepackage{amsmath,amssymb,amsfonts}
\usepackage{algorithmic}
\usepackage{graphicx}
\usepackage{textcomp}
\usepackage{xcolor}
\def\BibTeX{{\rm B\kern-.05em{\sc i\kern-.025em b}\kern-.08em
    T\kern-.1667em\lower.7ex\hbox{E}\kern-.125emX}}
\usepackage{lmodern}
\usepackage{graphicx}
\usepackage{times}

\usepackage{multicol}
\usepackage{multirow}
\usepackage[font=small]{caption}

\usepackage{algorithm}
\usepackage{algorithmic}
\usepackage{listings}
\usepackage{cite}

\usepackage{siunitx,booktabs}
\usepackage{comment}
\usepackage{url}
\usepackage[justification=centering]{caption}
\usepackage{subcaption}
\usepackage{hyperref}
\hypersetup{
    colorlinks=true,
    linkcolor=blue,
    filecolor=magenta,      
    urlcolor=cyan,
    }
\newcommand{\etal}{\textit{et al}. }
\begin{document}

\title{Comparison Of Deep Object Detectors On A New Vulnerable Pedestrian Dataset}

\author{\IEEEauthorblockN{Devansh Sharma}
\IEEEauthorblockA{\textit{Department of Computer Science} \\
\textit{Bowling Green State University}\\
Bowling Green, United States \\
sharmad@bgsu.edu}
\and
\IEEEauthorblockN{Tihitina Hade}
\IEEEauthorblockA{\textit{Department of Computer Science} \\
\textit{Bowling Green State University}\\
Bowling Green, United States \\
thade@bgsu.edu}
\and
\IEEEauthorblockN{Qing Tian}
\IEEEauthorblockA{\textit{Department of Computer Science} \\
\textit{Bowling Green State University}\\
Bowling Green, United States \\
qtian@bgsu.edu}
}

\maketitle

\begin{abstract}
Pedestrian safety is one primary concern in autonomous driving. The under-representation of vulnerable groups in today's pedestrian datasets points to an urgent need for a dataset of vulnerable road users. In order to help train comprehensive models and subsequently drive research to improve the accuracy of vulnerable pedestrian identification, we first introduce a new dataset for vulnerable pedestrian detection in this paper: the BG Vulnerable Pedestrian (BGVP) dataset. The dataset includes four classes, i.e., Children Without Disability, Elderly without Disability, With Disability, and Non-Vulnerable. This dataset consists of images collected from the public domain and manually-annotated bounding boxes. In addition, on the proposed dataset, we have trained and tested five classic or state-of-the-art object detection models, i.e., YOLOv4, YOLOv5, YOLOX, Faster R-CNN, and EfficientDet. Our results indicate that YOLOX and YOLOv4 perform the best on our dataset, YOLOv4 scoring 0.7999 and YOLOX scoring 0.7779 on the mAP 0.5 metric, while YOLOX outperforms YOLOv4 by 3.8 percent on the mAP 0.5:0.95 metric. Generally speaking, all five detectors do well predicting the With Disability class and perform poorly in the Elderly Without Disability class. YOLOX consistently outperforms all other detectors on the mAP (0.5:0.95) per class metric, obtaining 0.5644, 0.5242, 0.4781, and 0.6796 for Children Without Disability, Elderly Without Disability, Non-vulnerable, and With Disability, respectively. Our dataset and codes are available at \href{https://github.com/devvansh1997/BGVP}{https://github.com/devvansh1997/BGVP}.
\end{abstract}


\section{Introduction}
The need for new standards for pedestrian safety has grown in the era of autonomous vehicles and advanced driver assistance systems (ADAS). Since 2018, there have been 11 documented instances of Tesla Auto-pilot causing collisions, and there have been about 37 test car crashes involving Uber, which has resulted in four fatalities \cite{online2}. In the US, 104K instances of pedestrian-related non-fatal accidents were documented in 2020 \cite{online4}. People over the age of 65 made up 20 percent of all pedestrian fatalities in 2020, while 1 in 5 children under the age of 15 who died in collisions were pedestrians \cite{online4}. 61 percent of adult users, according to a Brookings Institution survey, say they feel ``uneasy" in autonomous vehicles \cite{online1}. Even with new standards of safety in autonomous vehicles, there is some dissonance in how the public perceives the dependability of these self-driving mechanisms. When a self-driving Uber car hit a pedestrian crossing the street in 2018, the incident resulted in the death of the victim. The vehicle failed to react swiftly enough to avoid the crash because the software had incorrectly labeled the pedestrian as a ``False-Positive" \cite{online3}. All these reports imply the need for well-trained object detectors that do not just cater to the average pedestrian, but also to the more vulnerable groups.

Visual detection has recently become crucial to numerous industries, including autonomous driving. However, deep visual detectors face the risk of being biased and performing imprecisely on susceptible groups of people since vulnerable road users are underrepresented in most, if not all, popular pedestrian datasets. Such bias against at-risk pedestrians has less to do with the model architecture than the data used for training.

``Pedestrian" usually refers to a person walking along a road or in a nearby area. Most datasets ignore the vulnerability of pedestrians. The fact that vulnerable pedestrians are typically less common on the road and have significant differences in size and motor abilities, many existing datasets are unable to identify their vulnerability and incorrectly categorize them. They frequently concentrate on the average pedestrian, failing to recognize the bias that will arise from using these datasets to train object detection frameworks. To properly handle a traffic situation involving such groups, it is equally, if not more, crucial for the object detectors to correct and effectively evaluate pedestrians, who are generally more vulnerable. In light of this, we make two contributions in this work. A vulnerable pedestrian dataset will be introduced as the initial contribution, giving the community the resources they need to work on and enhance the effectiveness of object detection algorithms for vulnerable pedestrians. Our dataset, which includes underrepresented vulnerable road users, can also be used as a supplement to current general pedestrian datasets. The second contribution is that we examine and benchmark the performance of numerous cutting-edge detectors on our new dataset, including YOLOv4\cite{bochkovskiy2020yolov4}, YOLOv5\cite{yolov5}, YOLOX\cite{ge2021yolox}, Faster R-CNN\cite{ren2015faster}, and EfficientDet\cite{tan2020efficientdet}. Some other works have also utilized and cited our dataset\cite{priisalu2023semantic} \cite{ye2023improved}.

\section{Related Work} \label{sec:relatedWork}

We have witnessed an increase in datasets focusing on particular categories of objects as a result of the increased use of object detection for automation and security. With the use of these datasets, researchers try to customize the model's performance so that it performs well on both, a wide range of object classes and on specific subsets of objects within these classes. Pedestrians are one such object that pertains to the autonomous driving industry, where the ability to detect various pedestrians is one key factor of robust self-driving software.

In \cite{kumar2020p}, Kumar \etal present the P-DESTRE dataset that uses UAV video surveillance data for pedestrian detection and tracking. Richly Annotated Pedestrian (RAP)\cite{li2016richly} is an extensive dataset that provides pedestrian data from an uncontrolled scene with various viewpoints of the same area. Aimed at object tracking, \cite{lambert2018tsukuba} provides pedestrian data comprised of pedestrians' and cyclists' trajectories. OpenPTDS\cite{song2018openptds} has a similar approach where they introduced a dataset created from real-life experiments. They target the problem of pedestrian detection from the viewpoint of safety and security around metropolitan buildings.

The Caltech Pedestrian Dataset, which was released in 2009, is regarded as one of the most popular pedestrian datasets to date. It included an annotation tool and a substantial amount of video footage with annotated frames. In addition to comparing models (popular at that time) on their dataset, it also supplied a fresh standard for testing object detection algorithms on pedestrian data \cite{dollar2009pedestrian}. In \cite{zhang2017citypersons}, along with a new pedestrian dataset, a new per-frame methodology was introduced to analyze scale and occlusion. Using the thermal channel provided a different perspective for pedestrian detection in dark environments and \cite{hwang2015multispectral} provided a dataset of color-thermal pair images. 

Existing works on vulnerable pedestrians focus on offering a system, algorithm, or model. Song \etal~explore safety advancements and offer an action character deduction and analysis module for video streams focusing on vulnerable pedestrians in \cite{song2018vulnerable}. Pedestrian-Oriented Forewarning System (POFS)\cite{liu2015pofs} is another example mechanism implemented using smartphone communication to detect a possible accident. Their paper identifies pedestrians who are distracted by their mobile phones as vulnerable pedestrians and presents the aforementioned system to alert them in an adaptive manner.

The majority of research in pedestrian detection either does not focus on vulnerable pedestrians or ignores the potential impact of the data. Vulnerable pedestrians do not receive the attention they deserve since the existing datasets concentrate on the ordinary pedestrian. The absence of a dataset focusing on vulnerable road users translates to biases against vulnerable people such as children, the elderly, and people with disability. In this paper, we try to tackle this issue by taking a data-centric solution. We introduce a new dataset focusing on vulnerable pedestrians or road users rather than the average pedestrians. The vulnerability that we are interested in is related to a road user's age and physical disability. We hope that this new benchmark dataset will stem further research in this particular field.

\begin{figure*}
  \centering
  \begin{tabular}{@{}c@{} @{}c@{}}
    \includegraphics[width=1.0\linewidth]{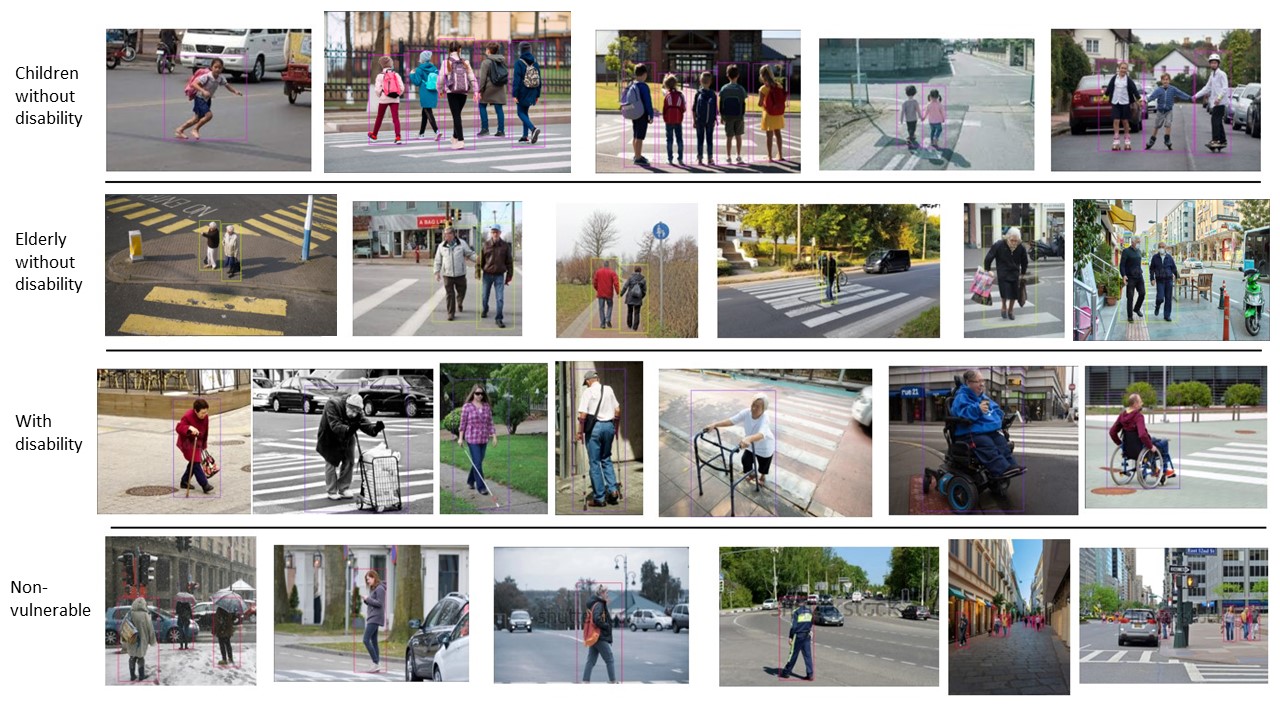} \\[\abovecaptionskip]
  \end{tabular}
    \caption{Simple traffic scenes in our BGVP dataset}\label{fig:allcls}
\end{figure*}

\section{BG Vulnerable Pedestrian (BGVP) dataset} \label{sec:dataset}
Vulnerable pedestrians deserve more attention from self-driving vehicles. It is currently not the case because of their under-representation in most training data, which results in unfair and discriminatory treatment of them.
Such a bias motivates us to create a dataset focusing on vulnerable pedestrians. There is a stark difference between an elderly disabled pedestrian and a young adult in their early 20s, crossing the road. Most accidents occur due to human errors, which may be avoided if autonomous driving becomes a standard. However, this requires exceptional functionality from the machines or software operating the vehicles. The object detectors should be able to quickly and precisely detect pedestrians and categorize them based on their vulnerability type. The vulnerability type is crucial information as it can help estimate the pedestrian's actions in a traffic scenario or emergency. The actions of a child aged 5-6, even when accompanied by an adult, can be considered unpredictable and risky. The detection of the child must be quick to control speed and predict the child's future positions with respect to the vehicle. This desirable behavior necessitates autonomous driving models to train/learn on a dataset where such vulnerable groups have a significant presence.

Over the years, many pedestrian datasets have been published, but none of them is dedicated to or mention the vulnerability of pedestrians \cite{zhang2017citypersons}\cite{kumar2020p}\cite{overett2008new}\cite{li2016richly}\cite{yang2019top}. We provide a new dataset that fills this gap and meets this critical demand. It will enable testing of future object detection models against it to determine how well a particular model can learn pedestrian vulnerability.  

We need to comprehend the many classes and how each pedestrian is qualified to fall into a certain class before delving into the dataset statistics. The bounding box instances in our dataset are divided into four categories: ``Children Without Disability", ``Elderly Without Disability", ``With Disability", and ``Non-vulnerable".
\begin{itemize}
\item \textbf{Children without Disability}: This vulnerable pedestrian class encompasses any pedestrian between 1 to 16 years of age. Most children are considered vulnerable because their behavior is potentially erratic, and they may not understand traffic laws. Some early teens can also be considered a risk to a self-driving car, which is why we include them in the same category. Children without a physical disability are considered to fall into this category. 

\item \textbf{Elderly without Disability}: Elderly pedestrians are particularly at risk. Even though the Elderly may be well versed with traffic laws, their age may restrict many physical motions, and on average, they tend to require more time to complete the same action compared to a non-vulnerable pedestrian. The minimum age of the senior age demographic is subjective and differs across cultures. In this paper, this group is determined by the first author to be people who appear in the age group of 50 or older. People from this age group without any physical disability are deemed eligible for this class.

\item \textbf{With Disability}: The most vulnerable group of all consists of pedestrians who have a physical impairment and may require some form of assistance. The pedestrian could be of any age group, but if they have a physical disability, they are categorized into this class because their age-related constraints are overshadowed by the disability. In our dataset, physical disability aids that we consider are wheelchairs, mobility walkers, scooters, crutches, and walking canes.

\item \textbf{Non-Vulnerable}: This group of pedestrians comprises of people that do not fall into any of the classes mentioned above, and it is safe to say that they are less vulnerable in traffic situations than other categories. These pedestrians do not have any physical (or visible) disability. They are supposed to act/respond quickly, and understand the situation better than vulnerable pedestrians. This is not to say that these pedestrians are not at all vulnerable; rather, their level of vulnerability in a traffic situation is lower than that of the categories mentioned above. 
\end{itemize}
\begin{figure*}
  \centering
  \begin{tabular}{@{}c@{} @{}c@{}}
    \includegraphics[width=1.0\linewidth]{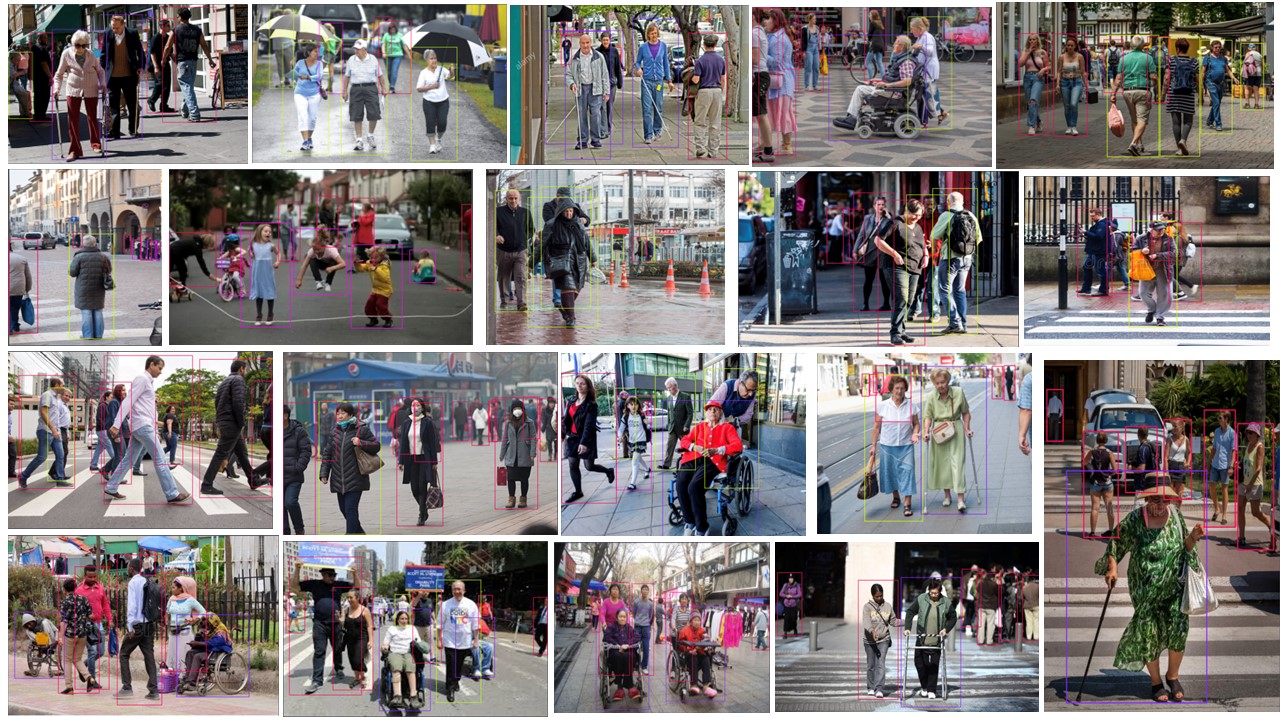} \\[\abovecaptionskip]
  \end{tabular}
    \caption{Complicated traffic scenes in our BGVP dataset}\label{fig:complscenes}
\end{figure*}

\begin{figure}
  \centering
  \begin{tabular}{@{}c@{} @{}c@{}}
    \includegraphics[width=1.0\linewidth, height=160pt]{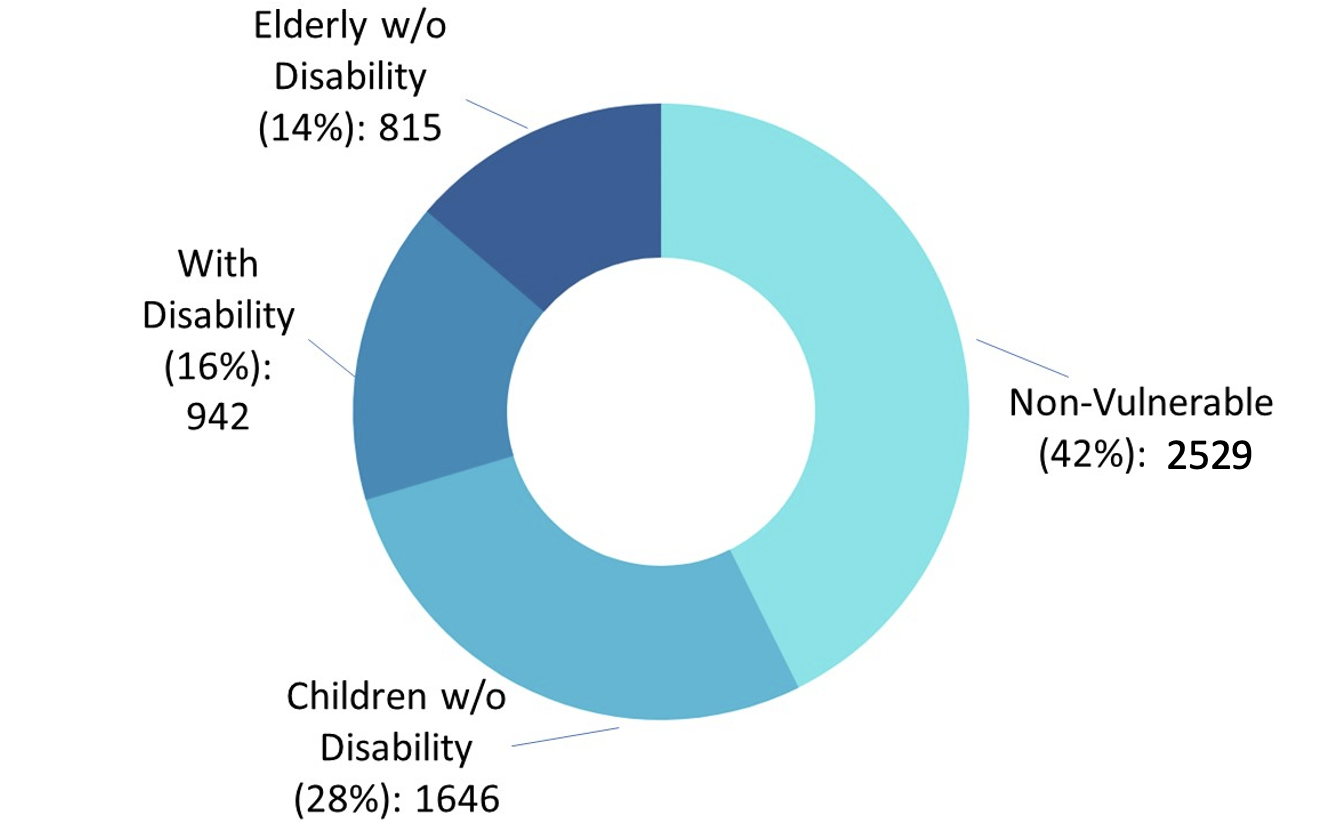} \\[\abovecaptionskip]
  \end{tabular}
    \caption{Annotations distribution among the four categories in our BGVP dataset}\label{fig:piechart}
\end{figure}

Figure \ref{fig:allcls} displays rows with a small sample of images from each class.
Figure \ref{fig:complscenes} displays a random selection of photos from our database that feature complex traffic scenes along with how our annotations appear on top of them \footnote{category labels are not shown for clarity}. Figure \ref{fig:complscenes} images have various class combinations, and some cover all classes.
Our dataset comprises 2,000 images with a total of 5,932 bounding box annotations. On average, there are three annotations in each image and the median image size is 600 $\times$ 408 pixels. The dataset contains an assortment of images with different dimensions. The dimension of the largest image is 6000 $\times$ 4000 and the smallest is 99 $\times$ 159. All images that we have collected are from the public domain (mainly the Internet). ``Children Without Disability” is the most prominent vulnerable pedestrian class, with 1,646 bounding box annotations, ``Elderly Without Disability” has 815 bounding box instances, ``With Disability” has 942 instances, and ``Non-vulnerable” has 2,529 instances. Fig. \ref{fig:piechart} visualizes the distribution of bounding box categories in the dataset. Our dataset is divided into 1,405 images as the training set, 399 images as the validation set, and 196 images as the testing set.

The vast majority of the photos we have gathered show these vulnerable people in traffic situations, which should aid in providing the detection models with pertinent contextual information which can benefit training. The contextual information typically includes traffic lights, zebra crossings, vehicles, and pedestrians (vulnerable or non-vulnerable).

It is also essential to find the right balance in the annotations that would help the model learn instead of confusing it even more, and our annotations would help in this regard. We annotate all pedestrians whose categories human eyes can easily recognize. If more annotations are created than necessary, that may lead to incorrect results. In our case, we have images that contain many pedestrians, but many in the background are too small in scale to be recognizable (we are unsure about what categories those pedestrians fall into). In such cases, we choose not to annotate them rather than label them incorrectly and mislead the detectors. 

\begin{table*}[h!]
\centering
\begin{tabular}{c c c c c c c } 
 \hline
Model & Backbone & Input Size & Params & GFLOPs & mAP 0.5 & mAP 0.5:0.95 \\ [0.5ex] 
 \hline\hline
 EfficientDet-D0 & Efficient-B0 & 512 & 3.9 M & 2.5 & 0.7048 & 0.4512 \\ 
 Faster R-CNN & Resnet-50 & 640 & 42 M & 180 & 0.7330 & 0.4860 \\
 YOLOv4 & CSPDarknet-53 & 640 & 27.6 M & 90.2 & 0.7999 & 0.5231\\
 YOLOv5-s & Modified CSP v5 & 640 & 7.2 M & 16.9 & 0.7000 & 0.4800 \\
 YOLOX-s & Modified CSP v5 & 640 & 8.94 M & 26.8 & 0.7779 & 0.5616 \\ [1ex] 
 \hline
\end{tabular}
\caption{Model Description and mAP Scores}
\label{table:1mainmap}
\end{table*}

\section{Models for Vulnerable Pedestrian Detection} \label{sec:models}
In this paper, we experiment with a variety of state-of-the-art object detection models on our introduced benchmark. We cover both single-stage and two-stage visual detectors, i.e., YOLOv4\cite{bochkovskiy2020yolov4}, YOLOv5\cite{yolov5}, YOLOX\cite{ge2021yolox}, EfficientDet\cite{tan2020efficientdet}, and Faster R-CNN\cite{ren2015faster}. 
\begin{table*}[h!]
\centering
\begin{tabular}{p{3cm} p{2cm} p{2cm} p{2cm} p{2cm}}
 \hline
 Model & Children Without Disability & Elderly Without Disability & Non-vulnerable & With Disability \\ [0.5ex] 
 \hline\hline
 EfficientDet-D0 & 0.4996 & 0.2932 & 0.3650 & 0.6469 \\ 
 Faster R-CNN & 0.5194 & 0.3889 & 0.3970 & 0.6371 \\
 YOLOv4 & 0.5360 & 0.4764 & 0.4494 & 0.6304 \\
 YOLOv5-s & 0.5120 & 0.3880 & 0.4040 & 0.6140 \\
 YOLOX-s & 0.5644 & 0.5242 & 0.4781 & 0.6796 \\ [1ex] 
 \hline
\end{tabular}
\caption{Per-class mAP 0.5:0.95 of the models}
\label{table:perclassmap}
\end{table*}

\subsection{Single-stage detectors}

Due to their high speed and efficiency, single-stage detectors are usually preferable in time-sensitive scenarios like autonomous driving. This is also the reason why most of the detectors tested in this paper belong to this category.

The most well-known one-stage detector family that has transformed the object detection community is the YOLO family. Redmon and his colleagues presented the first three versions of YOLO~\cite{Redmon_2016_CVPR, redmon2017yolo9000, redmon2018yolov3}, making important advancements along the way. YOLOv4~\cite{bochkovskiy2020yolov4} uses CSPDarknet53 backbone, SPP additional module and PANet path-aggregation neck. The detector head is similar to that of YOLOv3. YOLOv4 also introduces new data augmentation methods such as Mosaic and Self-Adversarial Training(SAT). YOLOv5 was released shortly after YOLOv4 by Ultralytics via GitHub\cite{yolov5}. YOLOv5 is similar to YOLOv4, but one difference is that YOLOv5 leverages auto-learning bounding box anchors. YOLOv5 also has different sizes, ranging from YOLOv5-S to YOLOv5-X, with S indicating the smallest width and depth while X denoting the largest. YOLOX~\cite{ge2021yolox} starts with YOLOv3 as its base and makes major changes, including (1) a decoupled head for classification, regression and localization, (2) dropping the use of anchors, (3) SimOTA advanced label assignment strategy, and (4) more advanced data augmentations techniques.

EfficientDet \cite{tan2020efficientdet}, developed by Google AI, is another single-stage detector, which is highly scalable and fast. EfficientDet uses EfficientNet\cite{tan2019efficientnet} as the backbone and a newly introduced BiFPN feature network. BiFPN enables easy and fast feature fusion and allows bi-directional information flow.

\subsection{Two-stage Detector}

Compared to single-stage detectors, two-stage detectors are usually slow due to the extra stage of regions of interest (RoI) proposal. Faster R-CNN~\cite{ren2015faster} is the most (at least one of the most) well-known two-stage detectors from Facebook AI Research (FAIR). It is faster than its predecessor Fast R-CNN \cite{girshick2015fast} because instead of passing all region proposals to the feature-extracting convnet, the convolution operation in Faster R-CNN is done only once per image to generate a whole feature map. The region proposal task is done by the proposed Region Proposal Network (RPN) in the latent feature space.

\section{Experiments and Discussion} \label{sec:experiments}

In this section, we benchmark the models previously mentioned on our introduced BGVP dataset and report their results. 

\subsection{Experimental Setup} 
We follow each model's authors' recommendations to set the hyper-parameters. We train the five state-of-the-art models on the proposed BGVP dataset. All models that are selected were pre-trained on MS-COCO dataset, allowing a fair comparison of their results on our dataset. Our training is performed on google colab and the Ohio Supercomputer Center (OSC) clusters. We report our results in terms of the Mean Average Precision (mAP). We use mAP 0.5 and mAP 0.5:0.95 as our primary metrics to analyze model performance. 
\begin{figure*}[ht]
\centering
 \begin{subfigure}[b]{0.32\linewidth}
 \includegraphics[width=\linewidth]{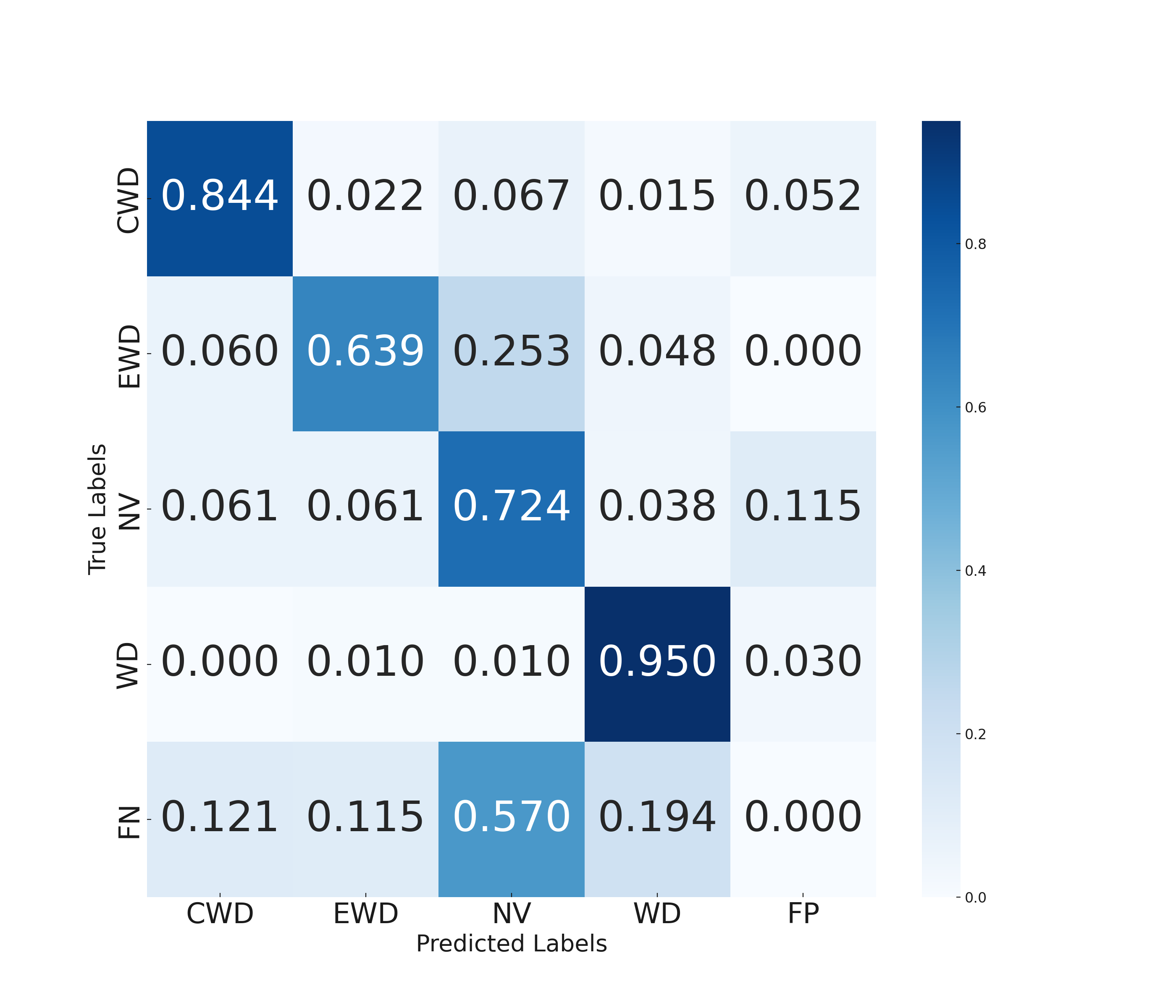} 
 \caption{YOLOX}
 \label{fig:yolox} 
 \end{subfigure} 
\quad 
 \begin{subfigure}[b]{0.32\linewidth}
  \includegraphics[width=\linewidth]{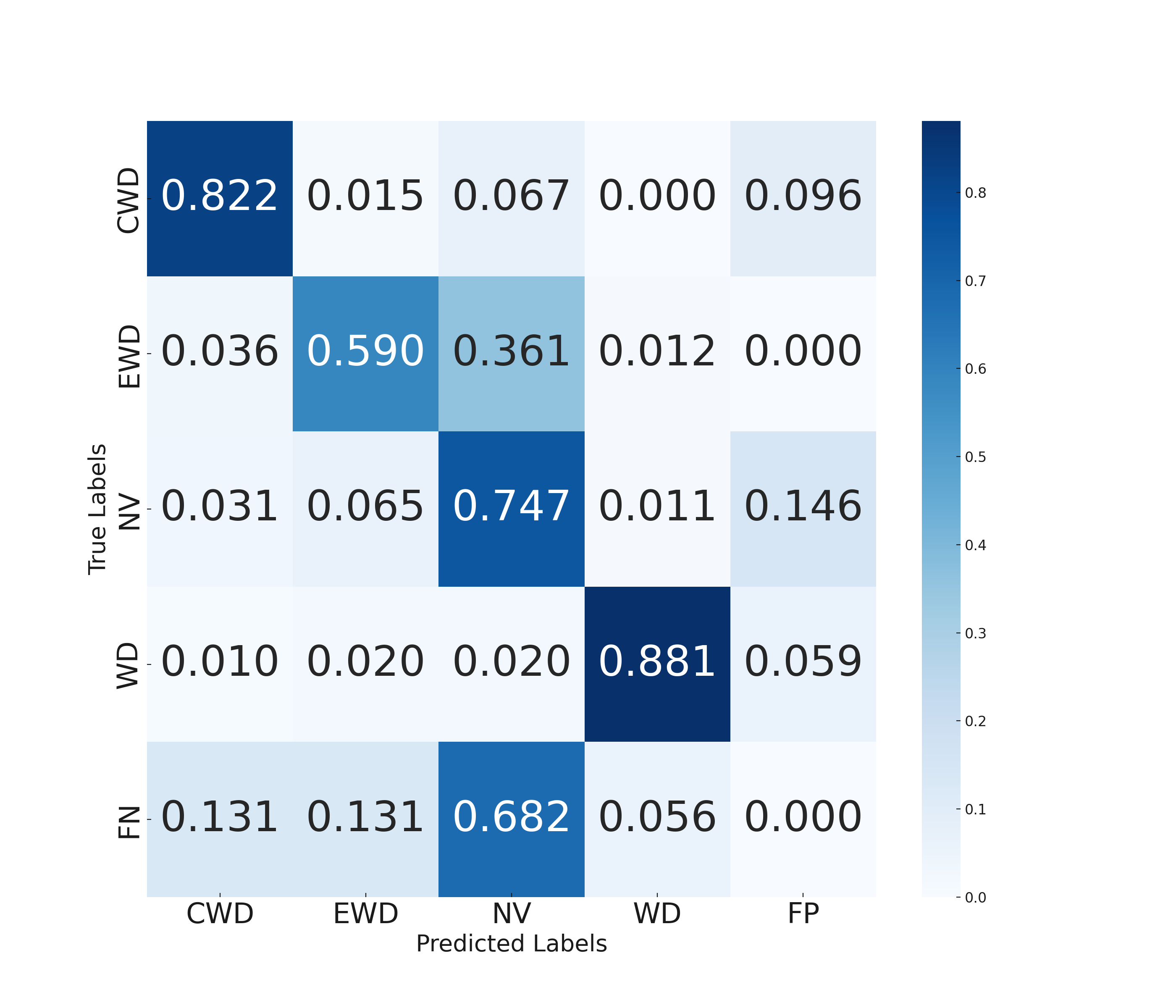} 
  \caption{YOLOv4}
  \label{fig:yolov4} 
 \end{subfigure}
 \begin{subfigure}[b]{0.32\linewidth}
  \includegraphics[width=\linewidth]{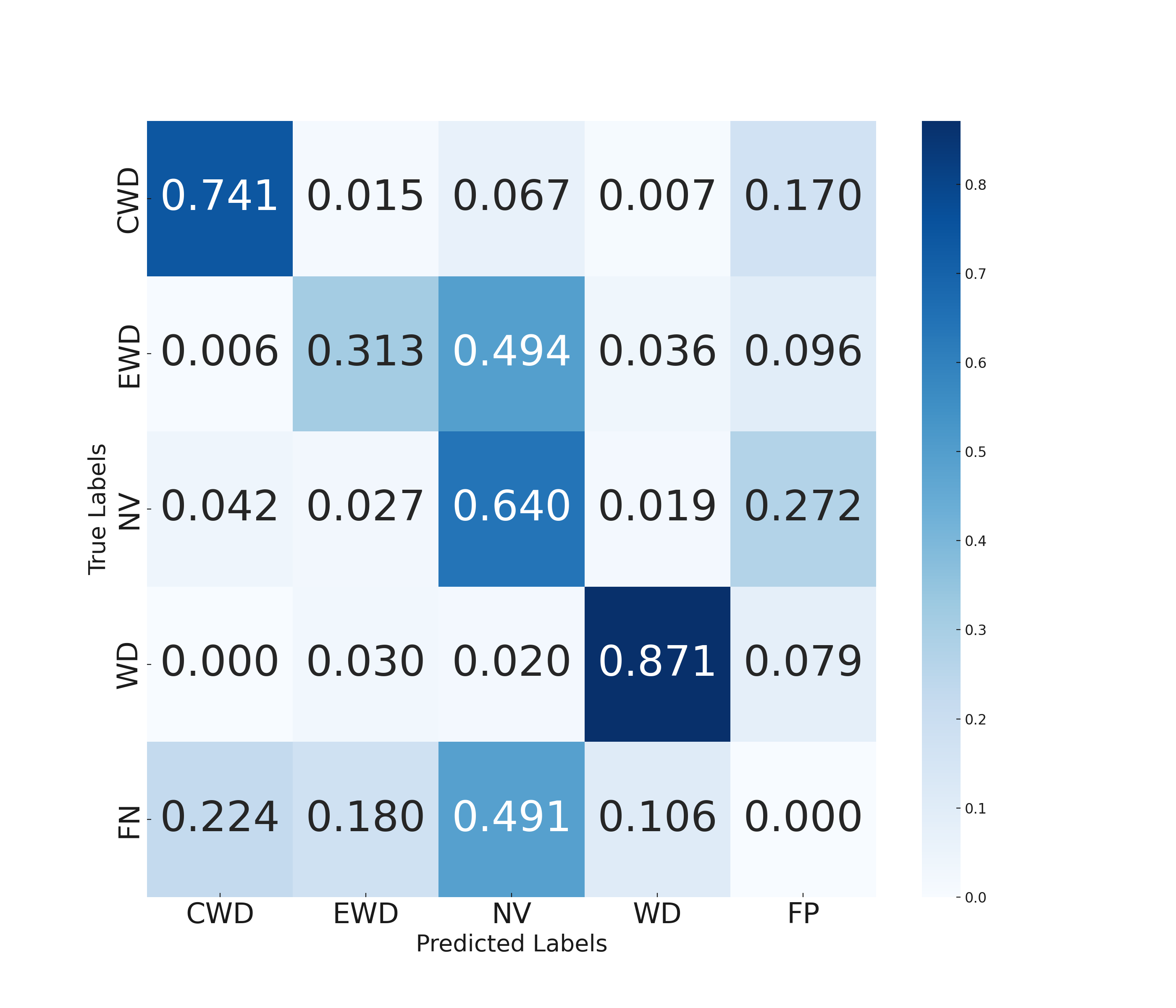} 
  \caption{EfficientDet}
  \label{fig:effdet} 
 \end{subfigure}
  \begin{subfigure}[b]{0.32\linewidth}
  \includegraphics[width=\linewidth]{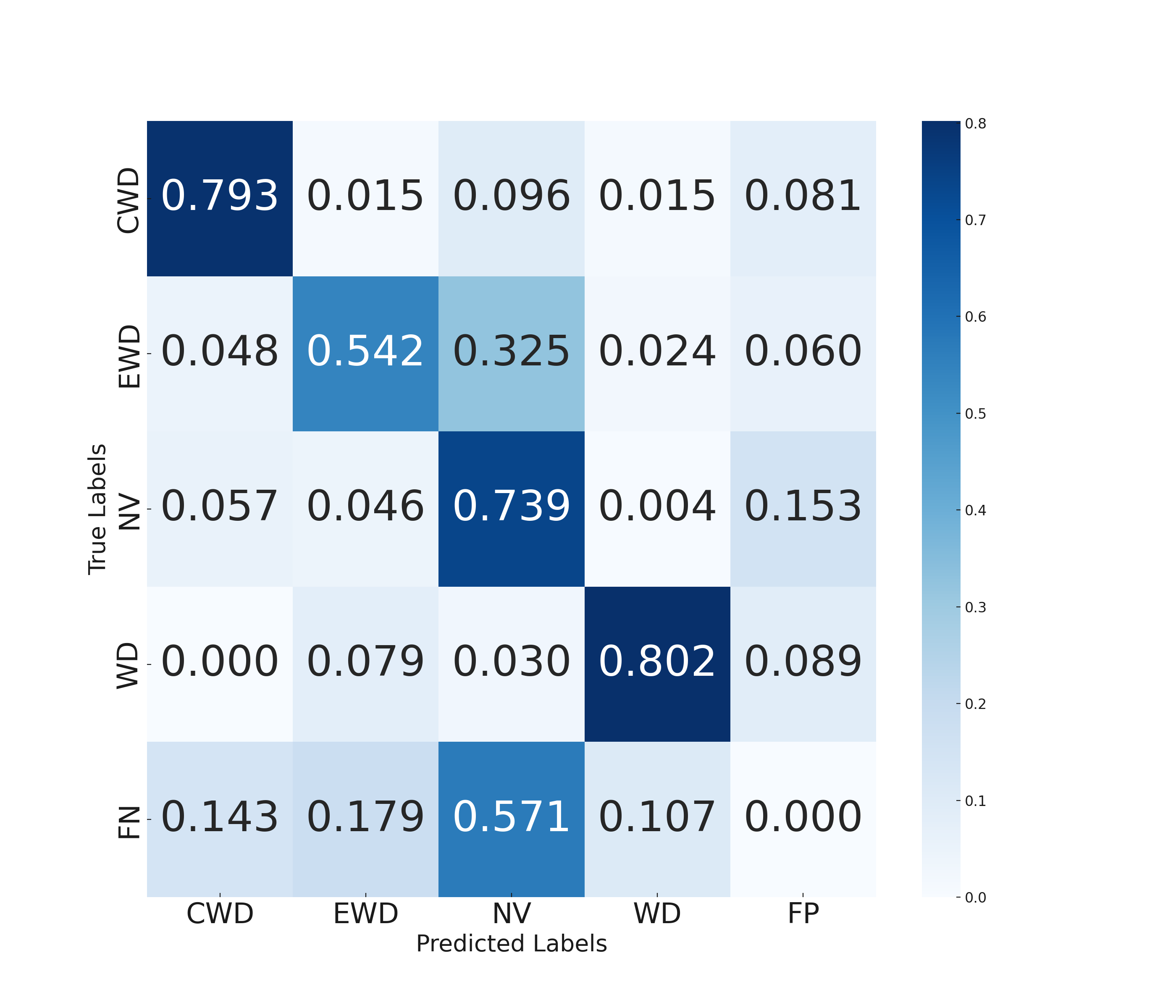} 
  \caption{Faster R-CNN}
  \label{fig:fasterrcnn} 
 \end{subfigure}
 \begin{subfigure}[b]{0.32\linewidth}
  \includegraphics[width=\linewidth]{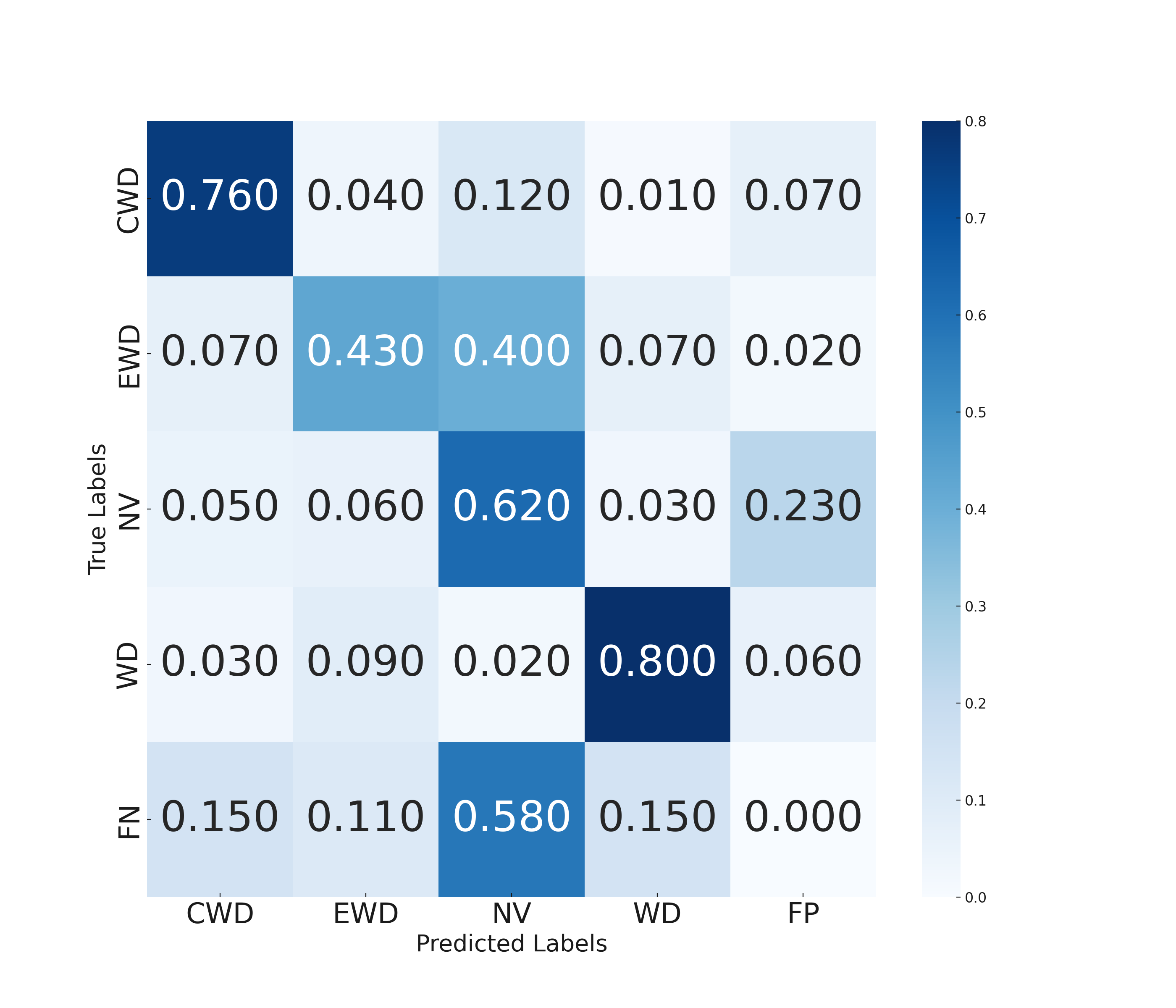} 
  \caption{YOLOv5}
  \label{fig:yolov5} 
 \end{subfigure}
\caption{Confusion Matrices of different models; CWD: Children Without Disability, EWD: Elderly Without Disability, NV: Non-Vulnerable, WD: With Disability, FP: False Positive, FN: False Negative
}
\label{fig:confmat}
\end{figure*} 
\subsection{Experimental Results}

Table \ref{table:1mainmap} shows the different models and their respective mAP scores. YOLOX and YOLOv4 perform the best among all. YOLOv4 has the highest score on mAP 0.5, scoring 0.7999 and YOLOX is closely behind scoring 0.7779. YOLOX achieved the best mAP 0.5:0.95 score (0.5616), outperforming YOLOv4 by 3.8 percent. Faster R-CNN, a two-stage detector, also performed reasonably well and scored 0.7330 for mAP 0.5 and 0.4860 for the mAP 0.5:0.95 metric.

In Table \ref{table:perclassmap}, we report per-class scores for the mAP 0.5:0.95 metric. It is evident that the majority of models, if not all, have higher mAP scores for the class ``With Disability". This could be due to the presence of regular forms that make up the physical walking aid, such as wheelchairs and walking canes. Relatively speaking, the models do not perform well in ``Elderly without Disability" and ``Non-vulnerable" classes. This is perhaps due to the unavoidable ambiguity and subjectivity when determining whether a subject meets the criteria for belonging to the senior age demographic. This uncertainty can also be seen in Figure \ref{fig:confmat}, which shows the confusion matrix results of all models. EfficientDet and YOLOv5 have a fair share of incorrect predictions of an elderly pedestrian as non-vulnerable. The YOLOX performs much better than other detectors in this regard. 
Performance on the class ``Children without Disability" seems more consistent across all models and averages to 0.5262 (mAP 0.5:0.95) while for the ``Elderly without Disability" class, the average is 0.4141. We also notice that all detectors perform poorly on images with a darker background. Many times they do not detect a pedestrian at all, irrespective of the category. 

\section{Conclusion}\label{sec:conclusion}

In this paper, we have presented a new object detection dataset, namely BGVP, for vulnerable pedestrians.
We have collected 2,000 images and have annotated 5,932 bounding box instances from four categories, i.e., ``Children Without Disability", ``Elderly without Disability", ``With Disability", and ``Non-Vulnerable". After collection and annotation, we trained and tested five state-of-the-art visual detectors and compared their results. We hope that this dataset can serve the community and motivate/facilitate more future research in this field. The dataset can also be utilized to fine-tune existing object detectors for more precise and less biased detection of vulnerable pedestrians, helping pave the way for safer and fairer autonomous driving.

\section*{Acknowledgment}

This work would not have been possible without the computing resources provided by the Ohio Supercomputer Center. Also, the authors would like to acknowledge Akinkunle A. Akinola for his help in this project. 


\end{document}